# AN INFORMATIONAL SPACE BASED SEMANTIC ANALYSIS FOR SCIENTIFIC TEXTS

Neslihan Suzen, Alexander N. Gorban,
Jeremy Levesley and Evgeny M. Mirkes

¹School of Computing and Mathematical Sciences,
University of Leicester, Leicester, UK

## ABSTRACT

*One major problem in Natural Language Processing is the automatic analysis and representation of human language. Human language is ambiguous and deeper understanding of semantics and creating human-to-machine interaction have required an effort in creating the schemes for act of communication and building common-sense knowledge bases for the 'meaning' in texts. This paper introduces computational methods for semantic analysis and the quantifying the meaning of short scientific texts. Computational methods extracting semantic feature are used to analyse the relations between texts of messages and 'representations of situations' for a newly created large collection of scientific texts, Leicester Scientific Corpus. The representation of scientific-specific meaning is standardised by replacing the situation representations, rather than psychological properties, with the vectors of some attributes: a list of scientific subject categories that the text belongs to. First, this paper introduces 'Meaning Space' in which the informational representation of the meaning is extracted from the occurrence of the word in texts across the scientific categories, i.e., the meaning of a word is represented by a vector of Relative Information Gain about the subject categories. Then, the meaning space is statistically analysed for Leicester Scientific Dictionary-Core and we investigate 'Principal Components of the Meaning' to describe the adequate dimensions of the meaning. The research in this paper conducts the base for the geometric representation of the meaning of texts.*

## 1. INTRODUCTION

One major problem in Natural Language Processing is the automatic analysis and representation of human language. Computational methods attempt to repeat human behaviour in the processing natural languages in a world where humans have no limitations on the range of interpretation of words, and the construction of complex meaning (semantic binding). Unlike humans as a group, machines may fail to provide a rich enough set of contexts to represent and distinguish different concepts.

The 'meaning of meaning' is a topic that has been extensively discussed by philosophers, linguistics, psychologists, neuroscientists, and computer scientists, in order to build "common-sense" knowledge bases, but the consensus has yet to be reached [1-4]. Wittgenstein formulates this as follows: "Meaning is use" or, in more detail, "For a large class of cases though not for all in which we employ the word 'meaning' it can be defined thus: the meaning of a word is its use in the language" [5, §43].

For the world of scientific texts (abstracts or brief reports), there is a well-defined dominant communicative function: a representative function. In an idealised scheme of the act of communication (see Figure 1), two representations of the situation on the "blackboards of consciousness" exist: the sender's representation (Representation 1) of the situation (Situation 1) and the receiver's representation (Representation 2) of the situation (Situation 2). A text related to the first situation is generated by the sender (Translation 1). This text is transmitted to the receiver and transformed by the receiver into a representation of the situation (Translation 2). The sender's and the receiver's representations never coincide.

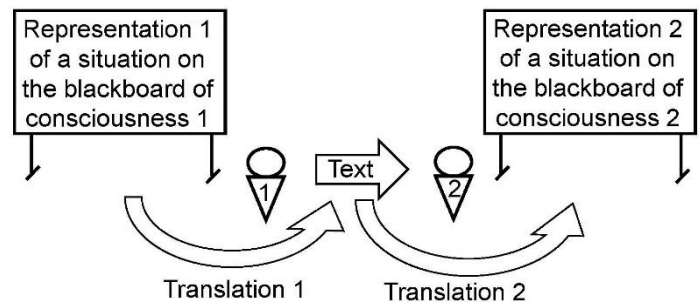

Figure 1. The idealised scheme of the act of communication. There is a representation of a situation on the sender's "blackboard of the consciousness" (a representation 1 of Situation 1). A text related to this situation is generated by the sender (Translation 1). This text is transmitted to the receiver and transformed by her into a representation of the situation (Representation 2 of Situation 2).

In this study, we consider the chain: Representation 1 → Text → Representation 2, and translations between them. Translations depend on a broad range of factors related to communication, including the experience of the sender and receiver. It is noteworthy that there may be many receivers and senders. One-to-many or even many-to-many communication add more situations and representations.

A very basic scheme is sufficient for our analysis of meaning. Meaning is hidden in the relationship between the representation of situations on the blackboard of the consciousness and the texts of the messages. The meaning of meaning can be understood if and only if the translation operations are created in the scheme of a communication act. Moreover, understanding can be represented as a reflexive game [6] with different levels (The sender prepares a message taking into account the experience of the receiver, his goals and tools, and guesses that the receiver takes into account the experience of the sender, her goals and tools, and... . Analogously, the receiver tries to understand the message taking into account..., etc.)

The relation between the text and the representation of the situation is a many-to-many correspondence. Each text corresponds to many situations and each situation can have many representative texts. At this stage, we characterise a situation "behind the text" by a set of attributes.

Despite the challenges in creating and describing plausible translation, with recent remarkable progress of machine translation, applying modern machine learning tools seems to be attractive idea for the analysis and simulation of translation operations. However, there is no generally accepted tools for working directly with representations of situations, and we cannot propose a general solution to this problem. Such a solution, perhaps, is impossible in a finite closed form despite much effort over many decades.

Our goal is more modest. We will provide computational analysis of the relations between texts of messages and representations of situations for a large collection of brief scientific texts. Such representations must be standardised, at least in part, and expressed in the form of diagrams, specially organized texts, or by other means. The simplest and universal approach is to replace the situation representations with vectors of attributes. Sentiment analysis provides many examples of such representations. We aim to provide another basic example that is specific to scientific texts: a list of scientific subject categories that the text belongs to.

In any text classification, subject categories can be chosen by humans or a computer system with an understanding of the text, but conflicts of understanding are possible and maybe inevitable. Even famous preprint servers (such as arXiv), moderators sometimes change the category selected by the authors. This is because the content of the text may differ from its meaning [7], a confusion which often occurs (just as understanding the situation behind the text is often confused with recognising the content of the text).

In our analysis of meaning, the starting point is the combination of the text with the list of the subject categories the text belongs to – definition of the attributes of the situation behind the text. The key idea of this approach goes back to the lexical approach of Sir Francis Galton, who selected the personality-descriptive terms and stated the problem of their interrelations for real persons. Following his idea, in classical psycholinguistic studies, a similar approach was used in publications [8-10]. Osgood, with co-workers, in the theory of the Semantic Differential, hypothesised a 3-dimensional semantic space to quantify connotative meanings concerning psychological and behavioural parameters [11, 12]. They used an approach for the extraction of three 'coordinates of meaning' from the evaluation of the 'affective meaning' of words (objects) by people. The semantic space was built by, in his words, 'three orthogonal bipolar dimensions': Evaluation (E), Potency (P) and Activity (A). Of course, the research started considering many different attributes and these three were extracted by factor analysis. These evaluations of a single object were related to some situations involving this single object, not just to an isolated abstract object. The people evaluated not the abstract 'terms' but psychologically meaningful situations behind these terms; these situations were the sources of 'affective meaning'.

For our world of scientific texts, we characterise the situation of use by a scientifically specific description – the research subject categories of the text. Quantifying the meaning in our research follows the road: Corpus of texts + categories $\rightarrow$ Meaning Space (MS) for words + Geometric representation of the meaning of texts.

In our analysis of meanings, the starting point is to combine the text with the list of the subject categories the text belongs to. These categories can intersect: a text can belong to several categories as texts can be assigned to more than one category. The categories evaluate the situation (the research area) related to the text as a whole, not as a result of the combination of the meaning of words. This holistic approach defines the general meaning of a word in short scientific texts as the information that the use of this word in texts carries about the categories to which these texts belong. More explicitly, we quantify meaning by using the Relative Information Gain (RIG) (see Equation 7) for a word in a category. To do this we require two attributes of text $d$ for a given word $w_j$ and a given category $c_k$, defined as:

$c_k(d)$: The text $d$ is in the category $c_k$: Attribute values are Yes ($c_k(d) = 1$) or No ($c_k(d) = 0$);
$w_j(d)$: The word is in the text: Attribute values are Yes ($w_j(d) = 1$) or No ($w_j(d) = 0$).

In this approach, the corpus of scientific texts is a probabilistic sample space (the space of equally probable elementary results, each of which is a random selection of text from the corpus). $RIG(c_k, w_j)$ measures the (normalized) information about the value of $c_k(d)$, which can be extracted from the value $w_j(d)$ (i.e. from observing or not observing the word $w_j$ in the text $d$) for a text $d$ from the corpus. By this, we identify the importance of the word for the corresponding category in terms of information gained when separating the corresponding category from its complement.

To follow our road, a triad is needed: texts, dictionary and multidimensional evaluation of the situation of use presented by the categories. In this research, short scientific texts are abstracts of research articles or proceeding papers. For the first element of the triad, the whole world of abstracts is narrowed to a sample: 1,673,350 texts from the *Leicester Scientific Corpus (LSC)* [13]. The meaning of a word extracted from the corpus is represented by a 252-dimensional vector of RIGs, in which each of the texts in the LSC is assigned to at least one of these 252 Web of Science (WoS) categories [14]. Thus, we use these simple 252 binary attributes for multidimensional evaluation of the text usage situation, where the second element of the triad is the *Leicester Scientific Dictionary-Core (LScDC)* [15].

Next, a vector space to represent a word's meanings has been introduced: the *Meaning Space*. In the Meaning Space, coordinates correspond to the subject categories. Each word $w_j$ in the dictionary is represented by the vector $\overrightarrow{RIG_j}$, of information gains for the word for each of the subject categories. These vectors are estimations of the meaning of words as to their importance in each of the research fields. hypotheses here are: if words have similar vectors, they tend to have similar meanings, and if texts have a similar distributions of word meanings – similar clouds of word vectors – then they tend to have similar meanings (often referred to as the Distributional Semantic Hypothesis). We demonstrate that RIG-based word ranking is much more useful than ranking based on raw word frequency in determining the science-specific meaning and importance of a word. The proposed model based on RIG is shown to have ability to stand out topic-specific words in subject categories.

Having represented each word in the Meaning Space, these representations can be used in many text analysis problems including the creation of a thesaurus such as the *Leicester Scientific Thesaurus (LScT)* [16]. The LScT contains the most informative 5,000 words in science; in formativeness is measured as the average RIGs of a word across categories.

This representation scheme is the basis of the computational analysis of the meaning of texts and will be used later for our holistic approach to the meaning of text: the text is considered as a collection of words, the meaning of the text is hidden in a situation of use, which is evaluated as a whole.

In this study, the hypothesis that lexical meaning in science can be represented in a lower dimensional space rather than the 252-dimensional Meaning Space is tested. Principal Component Analysis (PCA) is performed to reduce the dimensionality of the Meaning Space, in which points are the 5,000 words of LScT and dimensions are categories. We analyse the dimension of the Meaning Space and visualise words and categories in the space of principle components (PCs). We interpret the first five PCs by their coordinates. For each component, categories are divided into three groups: categories that positively and negatively correlated with the corresponding component, and categories having near zero values in the component. Topics in these groups are analysed. We then analyse the extreme topic groups at opposite ends of the PCs in order to describe the PCs. Finally, different selection criteria (Kaiser, Broken Stick, an empirical method based on multicollinearity control – PCA-CN) are used to reduce the dimensionality of the category space to 61, 16 and 13, respectively.

## 2. DATASET

Our new approach is applied to construct the Meaning Space on the basis of Leicester Scientific Corpus (LSC) and Leicester Scientific Dictionary-Core (LScDC) [13, 14]. The LSC is a scientific corpus of 1,673,350 abstracts and the LScDC is a scientific dictionary of 103,998 words extracted from the LSC. Each text in the LSC belongs to at least one of the 252 subject categories of Web of Science (WoS). Words in the LScDC will be represented by 252-dimensional vector in the Meaning Space.

Finally, a thesaurus of science is created by selecting the most informative words from the LScDC. The informativeness here was measured by the average RIGs in categories. We introduced the *Leicester Scientific Thesaurus (LScT)* where the most informative 5,000 words from the LScDC were included in [16]. These words are considered as the most meaningful words in science. Later we will use the LScT in the study of the representation of the meaning of texts.

## 3. AN INFORMATIONAL SPACE OF MEANING

In this section, we introduce our novel vector space model developed for quantifying the meaning of words. The architecture of the approach to estimating the word meaning for a large family of natural language scientific texts has discussed. The new approach to word meaning is applied to construct the Meaning Space based on the LSC and LScDC.

We introduce the *Meaning Space*, in which the meaning of a word is represented by a vector of RIGs about the subject categories that the text belongs to. We hypothesize that words have scientifically specific meaning in categories and the meaning can be estimated by information gains from the word to the category. 252 subject categories of WoS are used in construction of vectors of information gains. This representation technique is evaluated by analysing the top-ranked words in each category. For individual categories, RIG-based word ranking is compared with ranking based on raw word frequency in determining the science-specific meaning and importance of a word.

We finally create a scientific thesaurus, LScT, in which the most informative words are selected from the LScDC by their average RIGs in categories. LScT contains the most informative 5,000 words in the corpus LSC. These words are considered as the most meaningful words in science.

### 3.1. Word Meaning as a Vector of RIGs Extracted for Categories

We start with measuring how informative a word is for a category in terms of its ability to separate the corresponding category from its set theoretical complement. We hypothesize that topic-specific words in categories have larger information gain than other words, and such words are expected to have less gain in most other categories. Therefore, we approach this problem by defining, for each subject category $c_k$, a random Boolean variable: the text belongs to the category $c_k$ or the text does not belong to the category $c_k$ (this class is denoted as $\bar{c}_k$). The frequencies of words in classes $c_k$ and $\bar{c}_k$ are shown in Table 1. Let $D^j$ denote the set of texts containing the word $w_j$ and $D_k$ be the set of texts in the category $c_k$.

For every word $w_j$ from the dictionary (j = 1, ..., N) and every text $d_i$ from the corpus (i = 1, ..., M) the indicator $w_j(d_i)$ is defined as follows: If the word $w_j$ occurs in the text $d_i$ (once or more), then $w_j(d_i) = 1$, otherwise, $w_j(d_i) = 0$. The frequency of the word $w_j$ in the category $c_k$ is

$$w_{jk} = \sum_{d_i \in D_k} w_j(d_i);  \qquad (1)$$

$w_{jk}$ is the number of texts containing the word $w_j$ in the category $c_k$.

Table 1. Representation of the word by a pair of frequencies: the number of texts containing the word $w_j$ that belong and does not belong to the category $c_k$

| Word \ Category | $c_k$ | $\overline{c}_k$ |
|---|---|---|
| $w_1$ | $w_{1k}$ | $\lvert D^1 \rvert - w_{1k}$ |
| $w_2$ | $w_{2k}$ | $\lvert D^2 \rvert - w_{2k}$ |
| $\vdots$ | $\vdots$ | |
| $w_N$ | $w_{Nk}$ | $\lvert D^N \rvert - w_{Nk}$ |

Since words are obviously not mutually exclusive (one text usually contains several different words), to evaluate the information gain of the category $c_k$ from the word $w_j$ it is necessary to introduce, for each word $w_j$, a random Boolean variable with two states: $w_{jk}$ denotes the presence of the word in texts of the category $c_k$ and $\overline{w_{jk}}$ denotes the absence of the word $w_j$ in texts of the category $c_k$. The 2 × 2 contingency table used to calculate the information gain of the category $c_k$ from the word $w_j$ is presented in Table 2.

Table 2. Contingency table for the category $c_k$ and the word $w_j$

| Word \ Category | $c_k$ | $\overline{c}_k$ | Total |
|---|---|---|---|
| $w_j$ | $w_{jk}$ | $\lvert D^j \rvert - w_{jk}$ | $\lvert D^j \rvert$ |
| $\overline{w_j}$ | $\lvert D_k \rvert - w_{jk}$ | $M - \lvert D_k \rvert - (\lvert D^j \rvert - w_{jk})$ | $M - \lvert D^j \rvert$ |
| Total | $\lvert D_k \rvert$ | $M - \lvert D_k \rvert$ | M |

A general concept for computing information is the Shannon entropy [17]. *Information Gain (IG)* is a common feature selection criterion in machine learning used, in particular, for evaluation of word goodness [18, 19]. Information gain is a measure of the information extracted about one random variable if the value of another random variable is known. It is closely related to the mutual information that measures the statistical dependence between two random variables. The larger the value of the gain, the stronger the relationship between the variables.

The goal of this research is to evaluate the informativeness of words for category identification, and use this informativeness for word ranking and text representations. Therefore, we will consider information gain of the category $c_k$ from the word $w_j$: $IG(c_k, w_j)$. This information gain evaluates the number of bits extracted from the presence/absence of the word $w_j$ in the text for the prediction of this text belonging to category $c_k$. One may expect that if a word is a very topic-specific for a category, it appears in texts belonging to this category more frequently than in texts which do not belong to this category, and the majority of texts belonging to this category contain the word.

For each category, $c_k$, a function is defined on texts that takes the value 1, if the text belongs to the category $c_k$, and 0 otherwise. For each word, $w_j$, a function is defined on texts that takes the value 1 if the word $w_j$ belongs to the text, and 0 otherwise. We use for these functions the same notations $c_k$ and $w_j$. Consider the corpus as a probabilistic sample space (the space of equally probable elementary outcomes). For the Boolean random variables, $c_k$ and $w_j$, the joint probability distribution is defined according to Table 2. The entropy and information gains can be defined as follows.

The information gain about category $c_k$ from the word $w_j$, $IG(c_k, w_j)$, is the amount of information on belonging of a text from the corpus to the category $c_k$ from observing the word $w_j$ in the text. It can be calculated as [17]:

$$IG(c_k, w_j) = H(c_k) - H(c_k|w_j), \qquad (2)$$

where $H(c_k)$ is the Shannon entropy of $c_k$ and $H(c_k|w_j)$ is the conditional entropy of $c_k$ given the observing the word $w_j$. Entropies $H(c_k)$ and $H(c_k|w_j)$ are computed as follows:

$$H(c_k) = -P(c_k) \log_2 P(c_k) - P(\bar{c_k}) \log_2 P(\bar{c_k}). \qquad (3)$$

$P(c_k)$ is the probability that the text belongs to the category $c_k$, $P(\bar{c_k})$ is the probability that the text does not belong to the category $c_k$. Furthermore,

$$
\begin{aligned}
H(c_k|w_j) = & P(w_j)\big(-P(c_k|w_j) \log_2 P(c_k|w_j) - P(\bar{c_k}|w_j) \log_2 P(\bar{c_k}|w_j)\big) \\
& + P(\overline{w_j})\big(-P(c_k|\overline{w_j}) \log_2 P(c_k|\overline{w_j}) - P(\bar{c_k}|\overline{w_j}) \log_2 P(\bar{c_k}|\overline{w_j})\big),
\end{aligned} \qquad (4)
$$

where

- $P(w_j)$ is the probability that the word $w_j$ appears in a text from the corpus;
- $P(\overline{w_j})$ is the probability that the word $w_j$ does not appear in a text from the corpus;
- $P(c_k|w_j)$ is the probability that a text belongs to the category $c_k$ under the condition that it contains the word $w_j$;
- $P(\bar{c_k}|w_j)$ is the probability that a text does not belong to the category $c_k$ under the condition that it contains the word $w_j$;
- $P(c_k|\overline{w_j})$ is the probability that a text belongs to the category $c_k$ under the condition that it does not contain the word $w_j$;
- $P(\bar{c_k}|\overline{w_j})$ is the probability that a text does not belong to the category $c_k$ under the condition that it does not contain the word $w_j$.

All the required probabilities, entropies and relative entropies are evaluated using the contingency Table 2:

$$H(c_k) = -\frac{|D_k|}{M}\log_2\frac{|D_k|}{M} - \frac{M-|D_k|}{M}\log_2\frac{M-|D_k|}{M}, \qquad (5)$$

and

$$
\begin{aligned}
H(c_k|w_j) &= \frac{|D^j|}{M}\left(-\frac{w_{jk}}{|D^j|}\log_2\frac{w_{jk}}{|D^j|} - \frac{|D^j|-w_{jk}}{|D^j|}\log_2\frac{|D^j|-w_{jk}}{|D^j|}\right) \\
&+ \frac{M-|D^j|}{M}\left(-\frac{|D_k|-w_{jk}}{M-|D^j|}\log_2\frac{|D_k|-w_{jk}}{M-|D^j|}\right. \\
&\left. - \frac{M-|D_k|-(|D^j|-w_{jk})}{M-|D^j|}\log_2\frac{M-|D_k|-(|D^j|-w_{jk})}{M-|D^j|}\right).
\end{aligned}
\qquad (6)
$$

A high value of the informational gain $IG(c_k, w_j)$ (2) does not mean, in general, that the large proportion of information about a text belonging to the category $c_k$ can be extracted from observing the word $w_j$ in this text. This proportion depends on the value of the entropy $H(c_k)$ (5). The Relative Information Gain (RIG) measures this proportion directly. It provides a normalised measure of the Information Gain with regard to the entropy of $c_k$. RIG is defined as

$$RIG(c_k, w_j) = \frac{IG(c_k, w_j)}{H(c_k)}. \qquad (7)$$

We expect higher $RIG(c_k, w_j)$ for the topic-specific words in the category $c_k$. For simplicity, we denote $RIG(c_k, w_j) = RIG_{jk}$. Given the word $w_j$, $RIG_{jk}$ is used to form the vector $\overrightarrow{RIG_j}$, where each component of the vector corresponds to a category. Therefore, each word is represented by a vector of RIGs. It is obvious that the dimension of vector for each word is the number of categories K (for the WoS subject categories K = 252). For the word $w_j$, this vector is

$$\overrightarrow{RIG_j} = (RIG_{j1}, RIG_{j2}, \dots, RIG_{jK}). \qquad (8)$$

The set of vectors $\overrightarrow{RIG_j}$ can be used to form the *Word-Category RIG Matrix*, in which each column corresponds to a category $c_k$ and each row corresponds to a word $w_j$. Each component $RIG_{jk}$ corresponds to a pair $(c_k, w_j)$ and its value is the RIG from the word $w_j$ to the category $c_k$.

We define the Meaning Space as the vector space of such vectors $\overrightarrow{RIG_j}$. The dimension of this space is the number of categories and each coordinate is the RIG from a word to this category.

If we choose an arbitrary category, the words can be ordered by their RIGs from the most informative word to the least informative one. We expect that the topic-specific words will appear at the top of the list.

For a given word $w_j$, the sum $S_j$ of RIGs is calculated from the Word-Category RIG Matrix as:

$$S_j = \sum_{k=1}^{K} RIG_{jk}. \qquad (9)$$

The sum $S_j$ is a measure of the average informativeness of a word (this word has the informativeness $\frac{S_j}{K}$ on average). Now, the words in the dictionary can be ordered by their $S_j$. For each of these ordered lists of words, the most informative (meaningful) n words for scientific texts can be selected based on this criteria.

## 3.2. Experimental results

Having calculated RIGs for each word and created the Word-Category RIG Matrix, we evaluate the representation model by checking words in each category. That is, we consider the list of words with their RIGs in the corresponding category. Those words that have larger RIG are more informative in the category. Being 'more informative' here allows for the interpretation of being 'more specific' to the category's topic.

To visualise the top words in each category in a convenient way, we looked at word clouds. The font size of each word in a word cloud is proportional to its RIG in the category. For each category, words are sorted by their RIGs and the top 100 words are shown in the word clouds. Intuitively, the more informative the word is, the bigger size the word appears in word cloud Word clouds for the top 100 most informative words and histograms of RIGs for the top 10 most informative words for each of 252 categories can be found in [20].

In general, the RIG-based method proves to be more sensitive than the frequency-based method in identifying topic-specific words for a category. This means that representing words in Meaning Space has the advantage of transforming words to vectors efficiently with a benefit of considerably lower dimension than the standard word representation schemes.

To illustrate this result, we choose categories 'Biochemistry & Molecular Biology' and 'Mathematics' and compare two word clouds that are formed by using raw frequencies and RIGs in categories (see Figure 2 and Figure 3). It can be seen from the figures that the majority of the most frequent words in both categories are frequent words for the entire corpus. These words are not topic-specific for categories as they appear in almost all abstracts. The frequent but non-informative words can be seen  as generalised service words of Science and deserve special analysis. This proves that raw frequency is not important for identifying scientifically specific meanings of words. Therefore, by representing words as vector of RIGs, we can avoid such frequency bias. The most informative words in categories for RIG representation are topic-related in the corresponding category. We interpret these results as evidence for the usefulness of the RIG-based representation.

Words that are expected to be used together have very close values of RIGs. In " Health Care Sciences & Services", "health" and "care" are top words and RIGs for these words are so close (see Figure D.1 in [21]). Another example is "xrd" and "difract" in "Material Science, Ceramics". "XRD" is actually abbreviation of "X-ray diffraction"; therefore, they appear together as "X-ray diffraction (XRD)" for most of cases in the category (see Figure D.2 in [21]).

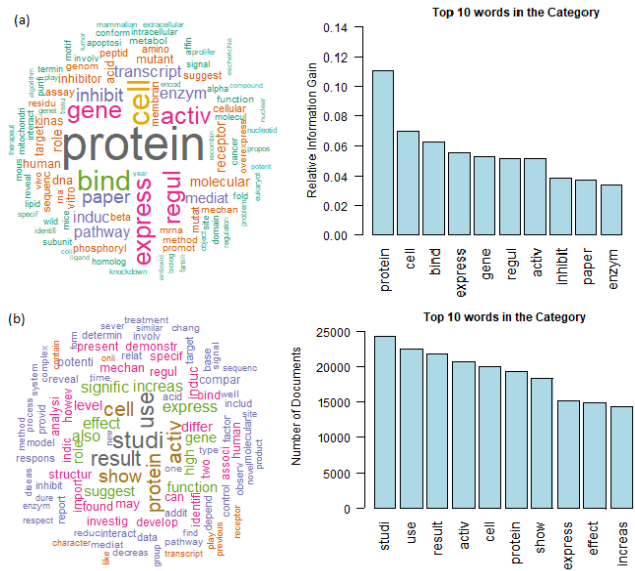

Figure 2. Category 'Biochemistry & Molecular Biology': word cloud of the top 100 most informative words and the histogram of the top 10 most informative words. The informativeness is defined by (a) RIG (b) frequency

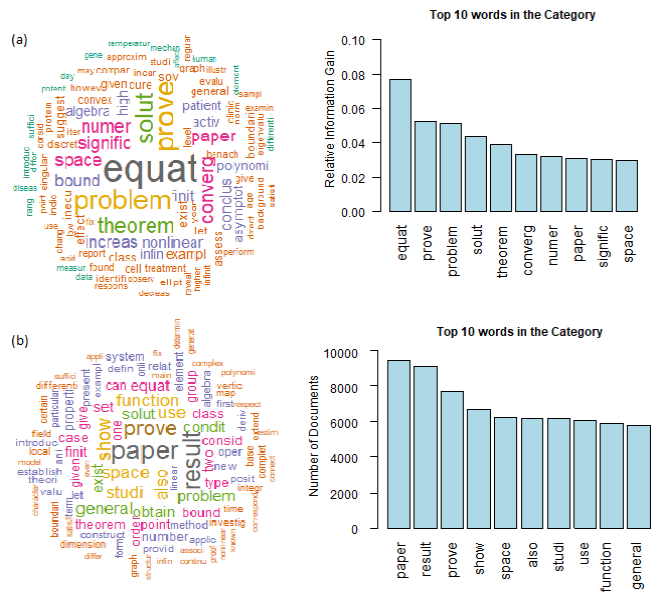

Figure 3. Category "Mathematics": word cloud of the top 100 most informative words and the histogram of the top 10 most informative words. The informativeness is defined by (a) RIG (b) frequency.

### 3.3. Thesaurus for Science: Leicester Scientific Thesaurus (LScT)

We have constructed the Word-Category RIG Matrix, where each entry corresponds to a pair (word, category) and its value shows the RIG for a text to belong to a category by observing this word in this text [21]. Row vectors of the matrix indicate the words' meaning in the scientific texts. A thesaurus of science was created by selecting the most informative words from the LScDC. The informativeness here was measured by the sum of RIGs in categories for this word.

We have introduced the Leicester Scientific Thesaurus (LScT): a list of 5,000 words which are created by arranging words of LScDC in their informativeness in the scientific corpus. The top 5,000 most informative words in the LScDC, where words are arranged by their $S_j$ are considered as the most meaningful 5,000 words in scientific texts. The full list of words in the LScT with their $S_j$ can be found in [16].

## 4. PRINCIPAL COMPONENTS OF MEANING

In this section, we hypothesize and test that lexical meaning in science can be represented in a lower dimensional space than 252. This space is constructed using PCA (singular value decomposition) on the matrix of word-category relative information gains. We argue that 13 dimensions is adequate to describe the meaning of scientific texts, and propose possibilities for the qualitative meaning of the principal components [22].

We apply PCA to reduce the dimensionality of the Meaning Space, in which points are 5,000 words of LScT and dimensions are categories. This section analyses the dimension of the Meaning Space and provides visualisation of words and categories in the space of PCs. In order to avoid redundant attributes in the data and identify the actual dimension of the space, we explore the Meaning Space by PCA.

We apply PCA and interpret the first five PCs by their coordinates (loadings). For each component, categories are divided into three groups defined as the main coordinates of the dimension and being unrelated attributes to the PC: categories that positively and negatively correlated with the corresponding component, and categories having near zero values in the component. We analyse the topics in these groups and visualise both categories and words on the PC axes. We also analyse the extreme topic groups at opposite ends of the PCs in order to describe the PCs based on extremely influential categories at both ends (10 categories at both ends).

Finally, by using three different selection criteria (Kaiser, Broken Stick, an empirical method based on multicollinearity control – PCA-CN), we reduce the dimensionality of the category space to 61, 16 and 13 respectively. Therefore, we argue that (lexical) meaning in science can be represented in a 13 dimension Meaning Space. We show that this reduced word set plausibly represents all texts in the corpus, so that the principal component analysis has some objective meaning with respect to the corpus. We argue that 13 dimensions is adequate to describe the meaning of scientific texts, and hypothesise about the qualitative meaning of the principal components.

### 4.1. Dimension of the Meaning Space

Given 252 subject categories, it is unreasonable to expect that every one of these categories is uncorrelated with all others (or distinct from them). For instance we might expect that the categories Literature and Literary Theory & Criticism will represent words in a very similar way in the Meaning Space (MS). Indeed, subcategories are likely to occur in the data and they are expected to have close values of RIGs for words. Such attributes will measure related information, and so the original 252 dimensional data contain measurements for redundant categories. Although the MS underlying the representation of word meaning has 252 dimensions, we expect that we will be able to represent words with significantly fewer dimensions in the MS.

An efficient way to represent words would be to map vectors onto a space that is constructed based on a combination of original features. Mathematically speaking, we look at a linear transformation from the original set of categories to a new space composed by new components. These new components are called *Components of the Meaning*. Two precise questions to be asked are: *how many components of meaning are there and how are these components constructed?* Thus, analysis of components (new attributes) based on the original attributes is crucial in understanding the MS. For instance, it is very important to understand which categories contribute the most and which the least to the new dimensions. Also, it is instructive to see if the new dimensions have some real semantic meaning, for instance, in distinguishing between natural and social sciences or experimental and theoretical research.

Words can be similarly represented in two or more categories. If two categories are correlated in the MS, it is possible to represent words in a reduced dimension by using a suitable linear combination of these original attributes. More specifically, if two categories are completely correlated, we would use the sum of two categories as one new attribute. The new attribute can be considered as a representative of the two original attributes. PCA provides a solution to this problem. Linear combination of weights (coefficients) is provided by PCA to create the new attribute, which we term a principal component (PC), with the aim of preserving as much variability as possible (the maximum variation in the data) [23,24]. The level of the effectiveness of PCA in explaining the data varies differently with the different sets of PCs. Therefore, in the sequel we investigate the effectiveness of PCA as a technique for determining the actual dimension of the data. Our goal is also to empirically investigate the effectiveness of the RIG-based word representation technique using PCs instead of the original attributes.

In PCA one of the crucial questions to answer is how many PCs should be selected. The Kaiser Rule is one of the methods developed to select the number of components [25, 26]. Eigenvalues of the covariance matrix are used to determine the appropriate number by taking components with eigenvalues greater than average of eigenvalues; only components explaining greater data variance than the original attributes should be kept [27].

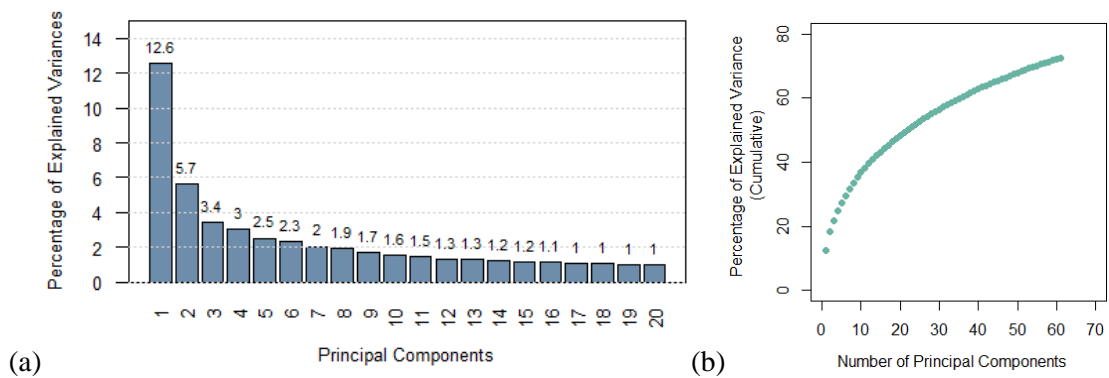

(a)                                                                                    (b)

Figure 4. (a) Fraction of variance explained as a function of PCs retained for categories; (b) Cumulative fraction of variance explained as a function of PCs retained for categories (61 PCs). The mean eigenvalue is 1.

PCs were assessed sequentially from the largest eigenvalue to the smallest. All PCs having eigenvalue less than average were considered to be trivial (non-significant) by the Kaiser rule. Hence 61 PCs are included as non-trivial, that is, 61 axes summarize the meaningful variation in the entire dataset. These non-trivial PCs are retained as informative at the first stage. The cumulative percentage of variance explained is displayed in Figure **4**. The cumulative percentage is approximately 73%, indicating the variance accounted for by the first 61 components. They explain nearly 73% of the variability in the original 252 attributes, so we can reduce the complexity of the data four times approximately, with only a 27% loss of information.

To interpret each component, the coefficients (influence) of the linear combination of the original attributes for the first five principal components are examined. The coordinates of the attribute divided by the square root of the eigenvalue gives the unit eigenvector, whose components give the cosine of the angle of rotation of the category to the PC. Furthermore, positive values indicate a positive correlation between an attribute and a PC and negative values indicate a negative correlation. Both the magnitude and direction of coefficients for the original attributes are taken into account. The larger the absolute value of the coefficient, the more important the corresponding attribute is in calculating the PC. Positive and negative scores in PCs push the overall score of a word in the meaning space to the right or left on the PC axis.

Following data reduction via PCA we then restricted the analysis of the informative categories to the non-trivial PCs; these are used to list informative attributes (categories). The importance of an attribute is determined as the maximum of the absolute values in coordinates of informative PCs for this attribute. The threshold $1/\sqrt{252}$ (threshold of importance) is used in the selection of informative attributes.

To examine the original attributes in the PCs, we introduce a threshold for categories having near zero values. The threshold used was $1/2\sqrt{252}$, which is half of the threshold of importance in selection of informative attributes. All values between $-1/2\sqrt{252}$ and $1/2\sqrt{252}$ are considered to be negligible so are in the zero interval. Hence, the initial attributes are considered as belonging to three groups: (1) positive, (2) negative, and (3) zero. We interpret the categories belonging to positive and negative groups as the main coordinates of the dimension as these categories contribute significantly to that direction. Categories belonging to the 'zero' group are seemed to be unrelated attributes to the PC. However, this information could be also useful. Hence, all categories in the three groups are meaningful and should be interpreted.

Categories in the three groups for each PC can be seen in Figures 3.3-3.7 in [22]. For the demonstration of the idea, we have displayed the second principal component in Figure 5. The zero interval is shown by a line in the figure. The number of categories in each group is presented in Table 3. The full list of categories in positive, negative and zero groups for each of five PCs can be found in Appendix C in [22].

We can see that there are no negative values for the first principal component. The first component primarily measures the magnitude of the contribution of categories to the PC. It is a weighted average of all initial attributes. The most prominent categories are 'Engineering, Multidisciplinary' and 'Engineering, Electrical & Electronic', that is, they strongly influence the component. This component explains 12.58 % of all the variation in the data. This means that more than 85% of the variation still retained in the other PCs.

Table 3. Number of categories in the groups of positive, zero and negative for the first five principal components

|  | PC1 | PC2 | PC3 | PC4 | PC5 |
|---|---|---|---|---|---|
| **Positive** | 221 | 63 | 60 | 67 | 55 |
| **Zero** | 31 | 131 | 131 | 129 | 142 |
| **Negative** | 0 | 58 | 61 | 56 | 55 |

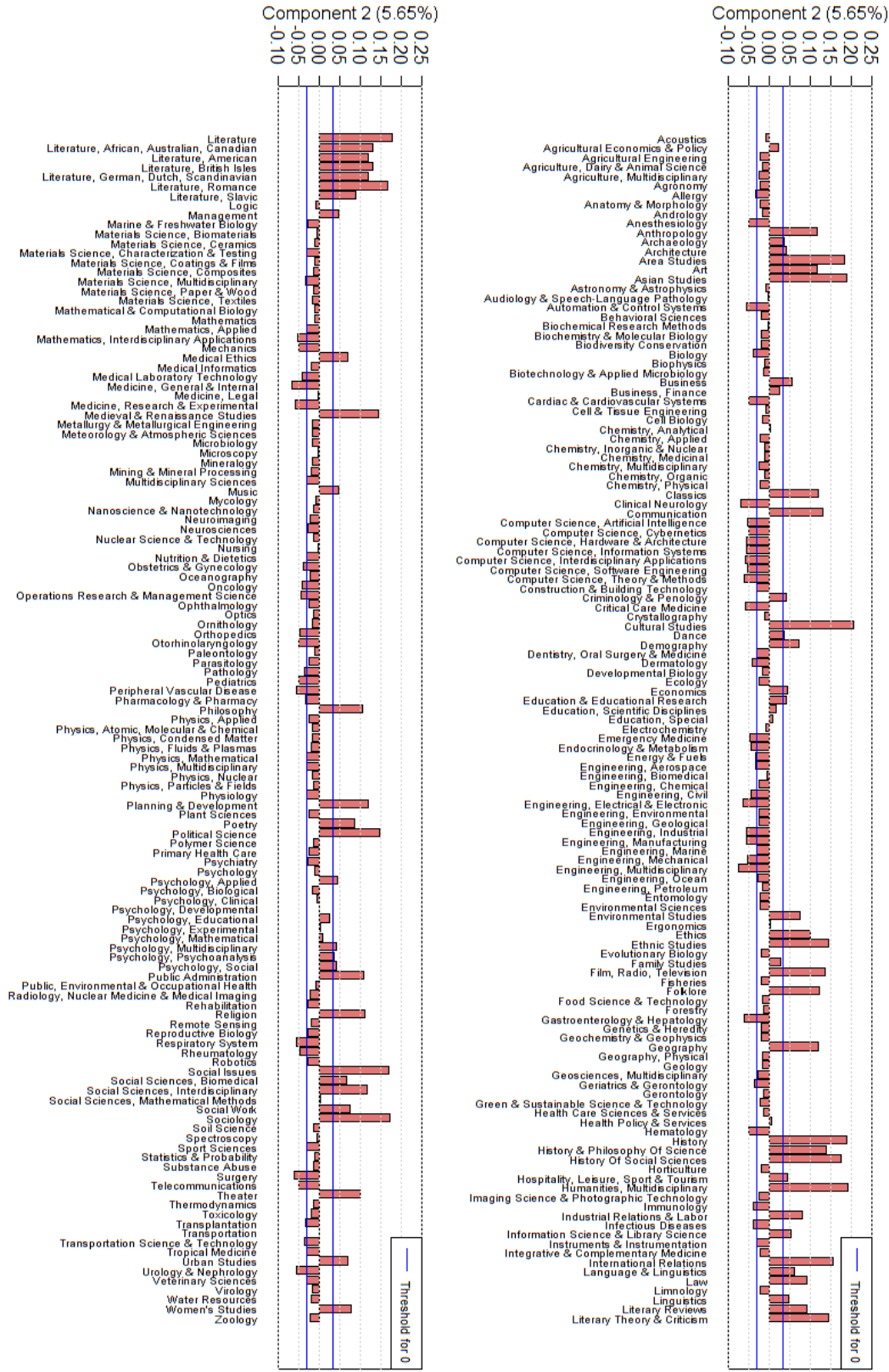

Figure 5. The second principal component of the LScT. The plot shows the contributions of original attributes (categories) on the second principal component.

The second component has positive associations with categories related to social sciences and humanities, and negative associations with categories related to engineering and natural sciences (see Figure 5). The plot shows that they are completely oppositely correlated. Hence, this component primarily measures the separation of two main branches of science. The most prominent category in the component is 'Cultural Studies'. The largest negative contribution to the component score is from the category 'Engineering, Multidisciplinary', which is approximately 2.5 times smaller than the contribution of 'Cultural Studies'. In the zero interval, extremely low values are present for attributes such as 'Psychology, Developmental', 'Ergonomics' and 'Medicine, Legal'.

The largest positive values on the third component can be interpreted as contrasting the biological science, computer science and engineering related areas with medicine, social care and some other disciplines. We may expect words that are used in biological science, computer science and engineering will go toward the positive side of the axis on the third principal coordinate. The largest negative values suggest a strong effect of psychology, medicine-health and physics related areas.

The other two principal components can be interpreted in the same manner. In the fourth component, the most prominent categories with positive values are some of social science branches such as economics, managements, psychology, ethics, education and multidisciplinary social science. Large negative values are for categories related to literature and medicine-health science. The fifth component has large positive associations with ecological, environmental sciences and geosciences.

We then analyse the topic groups at opposite ends of the PCs (positive and negative ends) in order to describe the PCs based on extremely influential categories at both ends. As such categories have high contributions in the PC, they are the parts of the trends in PCs and so explain the general trends of the PCs. This implies that we consider positive and negative groups introduced before, select the top $n$ categories with the highest component coefficients in each group and describe the grouping of categories in a way that categories at extreme ends can be distinguished from each other somehow and meaningfully described by a classification of research fields in science.

We implemented a heuristic technique. This approach starts with a search for a set of 10 categories with maximum coefficients at the two ends of the PC. The most informative 150 words are extracted in each of 10 categories, and the common words are listed. Words are analysed by human inspection to understand the meaning behind the opposite ends of the PC. The procedure is repeated for the PC2, PC3, PC4 and PC5. For the first PC1, the sign of coefficients are positive for all categories. High numbers for categories in this PC indicate that that category is well-described by words in the LScT.

The second PC seems to correspond a separation between discourse studies and experimental studies when we consider both the categories and words. For example, it is seen that three of the most informative common words are "argu", "polit" and "discours" for the groups of categories in the positive side and three of the most informative common words are "clinic", "treatment" and "therapi" for the groups of categories in the negative side in the PC. This is the **Nature of Science** dimension.

The third PC reflects two opposite types of research in terms of the requirement of microscopic and macroscopic instruments. At the positive end, scientific research mostly required detailed tools to work with the objects. Such tools can be instruments such as the microscope as well as programming tools used in coding. On the negative end, we are talking about human and population scale objects, but still related to humans. So this is the **Human Scale** dimension.

The fourth component appears to describe two classes of science: science of understanding the human condition through experiments and science of understanding the human condition through critical discourse studies. For instance, literary studies in the negative side are prominent and many texts from the literature are literary criticism of works. This is the **Human Condition** dimension.

Finally, the fifth component can be interpreted as contrasting natural science and intelligence. Categories related to natural science research are grouped in the positive extreme side and categories of understanding intelligence are located in the negative extreme side in this PC. 'Intelligence' can be both human intelligence and machine intelligence. For example, the categories 'Computer Science, Artificial Intelligence' and 'Psychology' are two of the top 10 categories. This is the **Inner World/Outer World** dimension.

### 4.2. Deciding the Dimension of the Meaning Space

The number of principal components determined by the Kaiser rule was 61. However, the Kaiser rule can underestimate or overestimate the number of PCs to be retained [28]. So, we also tested the Broken-Stick rule to determine the number of PCs [29-32]. Figure 6 demonstrates the optimal number of components determined by the Broken Stick and the Kaiser rules. The Broken Stick rule suggests that the reduction to only 16 PCs is reasonable.

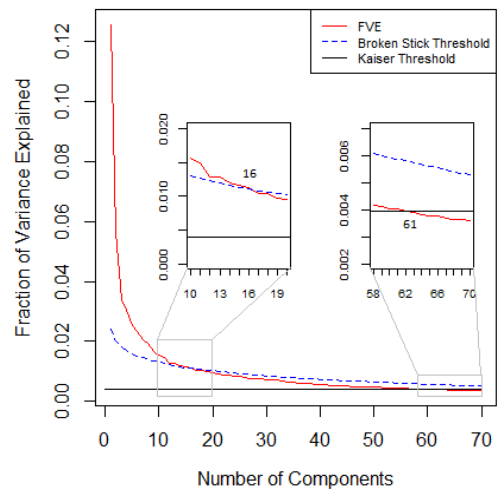

Figure 6. The number of principal components determined based on the Kaiser rule and the Broken Stick rule

Finally, we compared these two criteria of PC selection with the criterion: the ratio of the maximal and minimal retained eigenvalues ($\lambda_{max}/\lambda_{min}$) should not exceed the number of components selected (the condition number) [33-35]. This is described as multicollinearity control. To avoid the effects of multicollinearity, the conditional number of the covariance matrix after deleting the minor components should not be too large. That is, $k$ is the number of components to be retained if $k$ is the largest number for which $\lambda_1/\lambda_k < C$, where C is the conditional number. This method is called PCA-CN [35]. In our work, modest collinearity is defined using collinearity with $C = 10$ as in [34]. Therefore, the number of PCs to be retained is 13 by PCA-CN.

# 5. CONCLUSION AND DISCUSSION

In this work, we have initially studied the first stage of 'quantifying of meaning' for scientific texts: constructing the space of meaning. We have introduced the *Meaning Space* for scientific texts based on computational analysis of situations of the use of words. The situation of use of a word is described by the absence/presence of the word in the text in scientific subject categories. The meaning of the text is hidden in the situations of usage and should be extracted by evaluating the situation related to the text as a whole.

The situation of use is described by these 252 binary attributes of the text. These attributes have the form: a text is present (or not present) in a category. The meaning of a word is determined by categorising texts that contain the word and texts that do not. It is represented by the 252-dimensional vector of RIG about the categories that the text belongs to, which can be obtained from observing the word in the text.

We introduced an informational space of meaning for short scientific texts. The proposed word representation technique was developed and implemented on the basis of LSC with LScT. For concreteness, we followed the road: Corpus of texts + categories → Meaning Space for words. This involved the representation of words in the constructed Meaning Space and a detailed analysis of the Meaning Space. The proposed representation technique is evaluated by analysing the top-ranked words in each category. For individual categories, RIG-based word ranking is compared with ranking based on raw word frequency in determining the science-specific meaning and importance of a word.

We conclude that the use of informational semantics provides sizeable improvements to represent meaning in scientific texts over classical representation approaches based on raw frequencies, but how to make best use of it in different NLP tasks remains an open question that deserves further investigation.

This research has also introduced and analysed a scientific thesaurus LScT: a thesaurus of 5,000 words from the LSC. In the creation of the thesaurus, we have focused on the most informative words in science, which are the main scientific content words.

Our approach to meaning has been directed to meet some of the main challenges in extracting meaning from texts. First, it solves the problem of extracting the scientific-specific meanings because the proposed models of informational semantics characterise the situation of use by the subject categories of the text. Second, words have good representation for individual categories as well as the entire corpus because the relative importance of a word across all scientific categories is taken into account. Third, the creation of a space to represent words and texts is automated and reproducible so that it does not require a huge amount of human.

We also explore the Meaning Space by using Principal Component Analysis (PCA). We interpret the first five PCs by using their coordinates. We also suggest qualitative meanings for the first five of these dimensions. We welcome fierce debate over the meaning of these dimensions, but giving a qualitative meaning to these is a crucial step to understanding the meaning of meaning.
By exploring three different selection criteria (Kaiser, Broken Stick, PCA-CN) we reduced the dimensionality of the category space to 61, 16 and 13 respectively. If it turns out that we cannot explain some components of meaning at some time in the future with only 13 dimensions, we can increase the dimension. It remains a challenge to describe all 13 such dimensions in a way that makes some philosophical sense, but we hope that we have opened up this debate in this paper.

## AUTHORS

**Neslihan Suzen** (PhD) is Data Analytics and AI Innovation Fellow in the University of Leicester. She holds a PhD in the field of Natural Language Processing from the University of Leicester. Her research interests are focused on data analytics, machine learning and computational linguistics. She has practical hands-on experience in data science across a variety of fields.

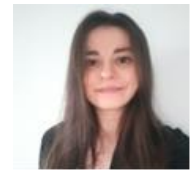

**Alexander N. Gorban** is a Professor in Applied Mathematics and the Director of the Centre for Artificial Intelligence, Data Analysis and Modelling (AIDAM) at the University of Leicester. He worked for Russian Academy of Sciences, Siberian Branch and ETH Zürich (Switzerland), was a visiting professor Clay Mathematics Institute (Cambridge, MA), IHES (Bures-sur-Yvette, France), Courant Institute of Mathematical Sciences (New York), and Isaac Newton Institute for Mathematical Sciences (Cambridge, UK). His main research interests are dynamical systems, biomathematics and machine learning.

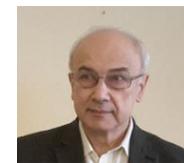

**Jeremy Levesley** (PhD, FIMA) is a Professor Emeritus in Applied Mathematics in the University of Leicester. He is a Senior Data Analyst at Redshift, University Liaison at Synoptix, and Senior Research Fellow at EMPAC. His research activity includes approximation in Euclidean space and on spheres using radial basis functions, and generalisations of these procedures to locally compact manifolds. He is interested in the applications of RBFs in finance, especially practical high dimensional approximation using sparse grid methods. Prof Jeremy is also interested in Smart City and Digital Medicine.

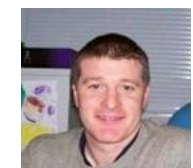

**Evgeny Mirkes** (Ph.D., Sc.D.) is a Research Associate at the University of Leicester, and a leader of the Data Mining group. His main research interests are biomathematics, data mining and software engineering, neural network and artificial intelligence. He has led and supervised many projects in data analysis and the development of decision-support systems for computational diagnosis and treatment planning and has participated in applied projects in Natural Language Processing in the area of social media data analysis. Dr Mirkes has rich experience in Predictive Mathematical and Computational Modelling and in finding solutions to classification, clustering, and auto coding problems.

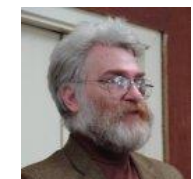